\documentclass[conference]{IEEEtran}
\usepackage{cite}
\usepackage{hyperref}
\hypersetup{
	hidelinks,
	colorlinks=true,
	linkcolor=red,
	citecolor=cyan,
	urlcolor = magenta
}
\usepackage{amsmath,amssymb,amsfonts}
\usepackage{algorithmic}
\usepackage{graphicx}
\usepackage{textcomp}
\usepackage{xcolor}
\usepackage{url}
\usepackage{pgfplots}
\usepackage{caption}
\usepackage{subcaption}
\usepackage{multirow}
\usepackage{booktabs}
\usepackage{adjustbox}
\usepackage{eso-pic}
\usepgfplotslibrary{groupplots,dateplot}
\renewcommand{\thetable}{\Roman{table}}
\def\BibTeX{{\rm B\kern-.05em{\sc i\kern-.025em b}\kern-.08em
    T\kern-.1667em\lower.7ex\hbox{E}\kern-.125emX}}
\AddToShipoutPictureBG*{
	\AtPageUpperLeft{
		\raisebox{-3\baselineskip}{
			\makebox[\paperwidth][c]{
				\parbox{0.9\textwidth}{\centering\footnotesize 
					© 2025 IEEE. Personal use of this material is permitted. Permission from IEEE must be obtained for all other uses, in any current or future media, including reprinting/republishing this material for advertising or promotional purposes, creating new collective works, for resale or redistribution to servers or lists, or reuse of any copyrighted component of this work in other works.
				}
			}
		}
	}
}
\usepackage{pgfplots, pgfplotstable}
\usepgfplotslibrary{groupplots}

\begin{document}

\title{TreeNet: A Light Weight Model for Low Bitrate Image Compression}

	\author{\IEEEauthorblockN{
		Mahadev Prasad Panda\IEEEauthorrefmark{2}, Purnachandra Rao Makkena\IEEEauthorrefmark{2}, Srivatsa Prativadibhayankaram\IEEEauthorrefmark{2}\IEEEauthorrefmark{4}, \\Siegfried Fößel\IEEEauthorrefmark{2}, Andr\'{e} Kaup\IEEEauthorrefmark{4}}
	\IEEEauthorblockA{
		\IEEEauthorrefmark{2}
		{ \textit{Moving Picture Technologies}, {Fraunhofer Institute for Integrated Circuits IIS, Erlangen}}\\
		\IEEEauthorrefmark{4}
		{ \textit{Multimedia Communications and Signal Processing}, Friedrich-Alexander-Universität Erlangen-Nürnberg, Erlangen}\\
		{ Germany}
}}
	
\maketitle

\begin{abstract}
Reducing computational complexity remains a critical challenge for the widespread adoption of learning-based image compression techniques. In this work, we propose TreeNet, a novel low-complexity image compression model that leverages a binary tree-structured encoder-decoder architecture to achieve efficient representation and reconstruction. We employ attentional feature fusion mechanism to effectively integrate features from multiple branches. We evaluate TreeNet on three widely used benchmark datasets and compare its performance against competing methods including JPEG AI, a recent standard in learning-based image compression. At low bitrates, TreeNet achieves an average improvement of 4.83\%  in Bj{\o}ntegaard delta bitrate over JPEG AI, while reducing model complexity by 87.82\%. 
Furthermore, we conduct extensive ablation studies to investigate the influence of various latent representations within TreeNet, offering deeper insights into the factors contributing to reconstruction.
\end{abstract}
\vspace{-0.15cm}
\section{Introduction}
\label{sec:intro}
With the proliferation of digital media, lossy image compression becomes vital.  
Traditional image compression methods such as JPEG2000 \cite{skodras2001jpeg}, BPG \cite{sullivan2012overview}, and VTM \cite{bross2021overview} have achieved impressive rate-distortion performance. However, these methods are constrained by their pronounced reliance on complex manually-engineered modules.  Recently, end to end optimized learned image compression has gained popularity.
The foundational learned compression methods \cite{balle2017endtoend, theis2017lossy} broadly follow the structure of a variational auto-encoder \cite{kingma2013auto}. The analysis transform $g_a$ transforms an image $x$ into a latent representation $y$. A quantizer $\mathcal{Q}$ quantizes $y$ into a discrete valued latent representation $\hat{y}$ followed by lossless entropy coding using an entropy model $\mathcal{E}_{\hat{y}}$. The synthesis transform $g_s$ maps the entropy decoded latent representation $\hat{y}_e$ to the image space to produce the reconstructed image $\hat{x}$. The overall process can be summarized as follows:
\begin{equation}
	y = g_a(x; \theta),\  \hat{y} = \mathcal{Q}(y),\  \hat{y}_e = \mathcal{E}_{\hat{y}} (\hat{y}),\  \hat{x} = g_s(\hat{y}_e; \phi)
\end{equation}
where, $\theta$ and $\phi$ represent learnable parameters of $g_a$ and $g_s$, respectively.
As quantization is non-differentiable, uniform noise is added during training to simulate the quantization process \cite{balle2017endtoend}. For optimization of such methods the following loss function is used
\begin{equation}
	\begin{split}
		\mathcal{L} &= \mathcal{R}(\hat{y}) + \lambda \cdot \mathcal{D}\left(x, \hat{x} \right),\\
	\end{split}
\end{equation}
where $\mathcal{R}(\hat{y})$ represents the bitrate of the discrete latent $\hat{y}$, and $\mathcal{D} \left(x, \hat{x}\right)$ indicates the distortion between the original image $x$ and the reconstructed image $\hat{x}$. The Lagrangian multiplier $\lambda$ controls the trade-off between rate and distortion. 
\begin{figure}[!t]
	\centering
	\includegraphics[width=0.49\textwidth]{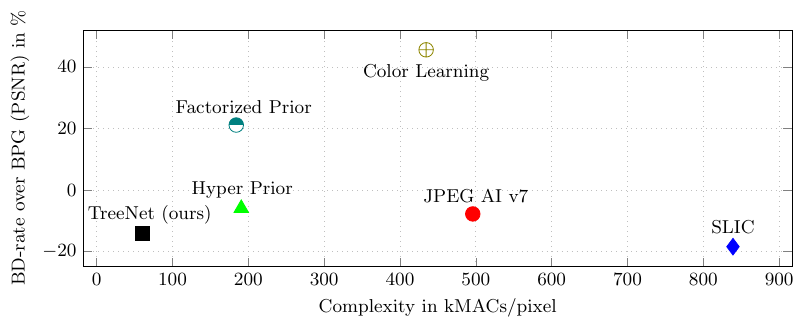}
	\includegraphics[width=0.49\textwidth]{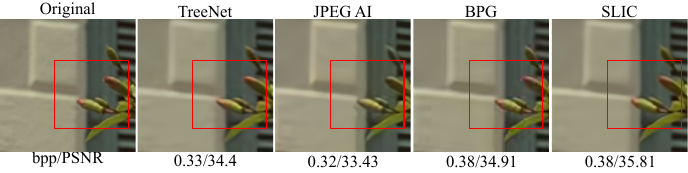}
	\vspace{-0.5cm}
	\caption{\textbf{Top:} BD-rate for PSNR on Tecnick dataset \cite{asuni2014testimages}. TreeNet achieves better trade-off between coding performance and complexity.  \textbf{Bottom:} Comparison of decoded patches on \emph{kodim06} image from Kodak dataset\cite{kodak1993}.}
	\label{fig:imgs}
	\vspace{-0.7cm}
\end{figure}

Several works concentrating on different facets of learned image compression, such as network design \cite{cheng2020learned, slic2024}, context modeling \cite{minen2018},\cite{ minnen2020channel},\cite{he2021checkerboard},\cite{he2022elic},\cite{jiang2023mlic},\cite{ jiang2023mlicpp} etc. have been proposed to enhance rate-distortion performance. Remarkably, some learned image compression methods \cite{he2022elic, jiang2023mlic, jiang2023mlicpp} are outperforming state-of-the-art traditional methods like the intra-coding mode of VVC. Recently, the JPEG standardization committee has introduced JPEG AI\cite{jpegai}, a learning-based image coding standard. However, to achieve better rate-distortion performance, these methods often necessitate complex model architectures, which limits their widespread adoption.
\begin{figure*}[t!]
	\begin{center}
		\includegraphics[width=0.9\linewidth]{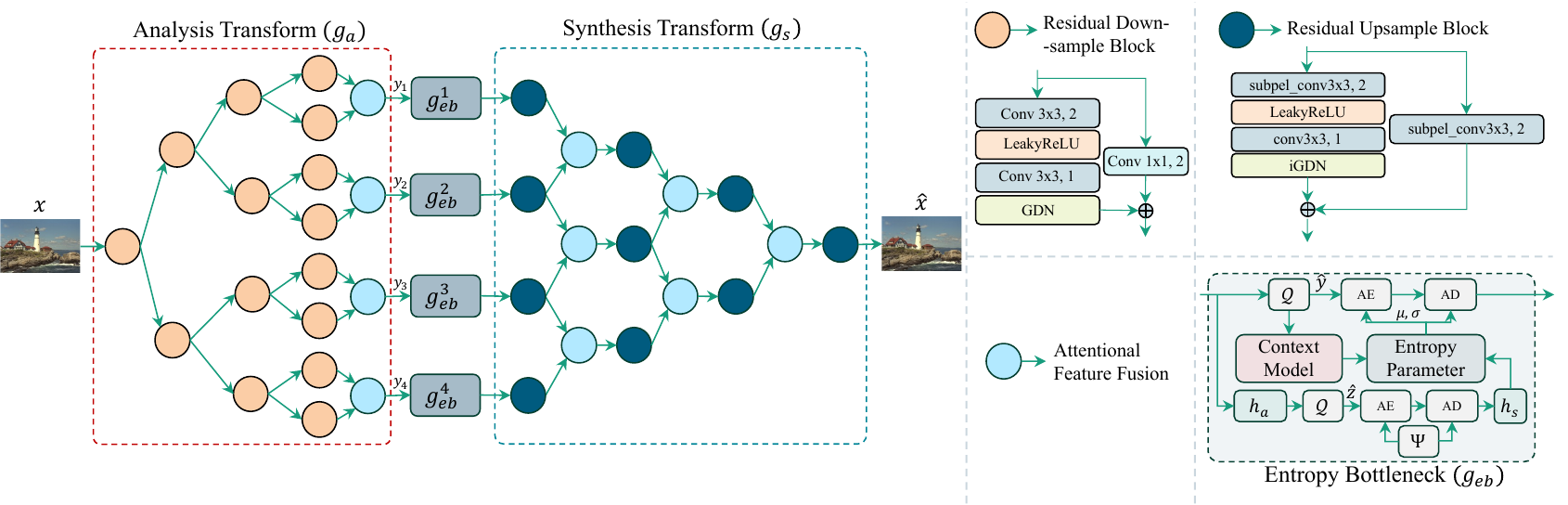}
	\end{center}
	\vspace{-0.3cm}
	\caption{Schematic diagram for TreeNet. A binary tree-based analysis transform $g_a$ maps image $x$ into four latents $y_1, y_2, y_3,$ and $y_4$. These latents are independently quantized and entropy coded using four entropy bottleneck blocks $g_{eb}^1, g_{eb}^2, g_{eb}^3$, and $g_{eb}^4$. The synthesis transform $g_s$ maps latents to image space producing reconstructed image $\hat{x}$. The attentional feature fusion is as described in \cite{dai21aff}. 
	}
	\label{fig:blockdiagram}
	\vspace{-0.5cm}
\end{figure*}
Various approaches have been introduced to address the computational cost associated with learned image compression. In \cite{minnen2020channel, he2021checkerboard} faster context models are proposed to accelerate encoding and decoding processes. Transformer-based architectures \cite{mentzer2023m2t, zhu2022transformerbased} have also been investigated to enhance computational efficiency, primarily through the parallel processing of independent sub-tensors. Highly asymmetric encoder-decoder architectures where the decoder is significantly less complex than the encoder have been proposed in \cite{yang2023computationally, galpinAssym}; though, the overall complexity remains high. Low complexity overfitted neural networks based on implicit neural representations have also been investigated for image compression \cite{leguay2023low, kim2024c3},\cite{ blard2024overfitted}. However, the time required for overfitting such networks poses a critical challenge. 

In contrast to these approaches, our work focuses on designing a computationlly less complex architecture for both the encoder and the decoder. Moreover, the proposed method is amenable to further optimization through parallel processing. With this context, we introduce TreeNet, a learning-based image compression model that takes advantage of a binary tree-structured encoder and decoder architecture. Such tree-structured network has previously been explored  for image denoising applications in\cite{flepp2024real}. 
Unlike the model proposed in\cite{flepp2024real},  our approach does not involve splitting of tensors. Furthermore, we incorporate attention-based feature fusion\cite{dai21aff} modules to combine features from different branches. 
Concretely, our contributions are summarized as follows,
\vspace{-0.0073cm}
\begin{itemize}
	\item We introduce a novel and computationally efficient learned image compression model, TreeNet, featuring a tree-based encoder and decoder design. 
	\item We provide a thorough evaluation of TreeNet, including both qualitative and quantitative performance analyses, as well as a detailed assessment of the model complexity.
	\item Through extensive experimentation, we elucidate the influence of various latent representations in TreeNet on the reconstructed image, thereby making our model more interpretable.
\end{itemize}
\vspace{-0.22cm}
\section{Model Architecture}
\label{sec:method}
In this section, we describe TreeNet in detail. Fig.~\ref{fig:blockdiagram} illustrates the schematic representation of the model architecture. The model consists an analysis transform $g_a(\cdot)$, four entropy bottlenecks $g_{eb}(\cdot)$, and a synthesis transform $g_s(\cdot)$.
\begin{figure}[t!]
	\begin{center}
		\includegraphics[width=\linewidth]{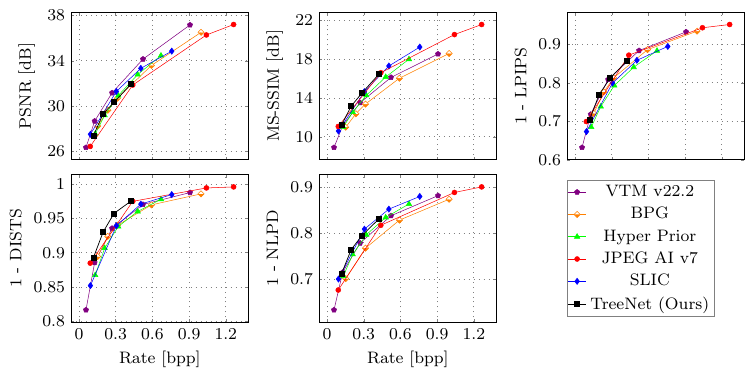}
		\vspace{-0.5cm}
		\caption{RD curves on Kodak dataset\cite{kodak1993}.}
		\label{fig:rdplot_kodak}
		\vspace{-0.8cm}
	\end{center}
\end{figure}
The analysis transform $g_a$ is designed as a perfect binary tree with a height of 3 encompassing 8 leaf nodes, each functioning as a learnable block. In our experiments, the nodes in $g_a$ are structured as residual downsampling blocks as depicted in Fig.~\ref{fig:blockdiagram}. The input image $x$ is processed by the root node of $g_a$. The preceding nodes form the input to the successive nodes, i.e., the two child nodes of each parent node receive identical input feature maps. This is done so that during training, the two child nodes have the scope of capturing unique and complimentary features from the complete input. 
We combine the feature maps emerging from the leaf nodes with common parent nodes using attentional feature fusion \cite{dai21aff}.  This results in four latents $y_1, y_2, y_3, $ and  $y_4$ which are then sent to four different entropy bottlenecks for latent specific entropy coding. 

The architecture of each entropy bottleneck is same as in\cite{cheng2020learned} consisting a hyper analysis transform $h_a(\cdot)$, a context model $g_c(\cdot)$, a hyper synthesis transform $h_s(\cdot)$, and an entropy parameter estimation block $h_{ep}(\cdot)$. However, instead of an autoregressive context model, we utilize the checkerboard context model\cite{he2021checkerboard} for faster inference. 

The output of four entropy bottlenecks is fed into the synthesis transform $g_s$. The synthesis transform consists of three upsampling layers. Each upsampling layer contains $N$ residual upsample nodes and $N-1$ feature fusion nodes, where $N \in \{3, 2, 1\}$. In each feature fusion node, the outputs from two residual upsample nodes are combined, as illustrated in Fig.~\ref{fig:blockdiagram}.  
In the end, a single residual upsample node generates the reconstructed image $\hat{x}$ utilizing the features received from the preceding upsampling layer. Existing model architectures\cite{ballé2018variational, cheng2020learned, minen2018, he2022elic} use 128 or 192 channels in the convolutional layers. However, TreeNet is configured with 32 channels, which reduces the complexity futher.
\vspace{-0.1cm}
\section{Experiments}
\label{sec:exp}
\subsection{Training Setup}
We trained the models on the combined train and test set of COCO 2017 dataset~\cite{lin2014microsoft} consisting 150,000 images. We randomly cropped patches of size $256 \times 256$ from the original images for training. The models are trained with a batch size of $16$ with random shuffling. 
For training the models following loss function was utilized,
\begin{equation}
	\small
	\begin{split}
		\mathcal{L} &= \mathcal{R}\left(\hat{y}, \hat{z}\right) + \lambda \cdot \mathcal{D}(x, \hat{x}) \\
		&= \mathbb{E}_{x \sim p_x}\left[-\sum_{i=0}^{3} \left(\text{log}_\text{2}(\mathcal{E}^{i}_{(\hat{y}_i|\hat{z}_i)}(\hat{y}_i|\hat{z}_i)) + \text{log}_\text{2}(\mathcal{E}^{i}_{(\hat{z}_i|\Psi)}(\hat{z}_i|\Psi))\right)\right] \\
		&+ \lambda_1 \cdot \mathbb{E}_{x \sim p_x} \left[MSE \left(x, \hat{x}\right)\right] \\
		& + \lambda_2 \cdot \mathbb{E}_{x \sim p_x} \left[1 -  \mathit{MS\mbox{-}SSIM}\left(x, \hat{x}\right)\right] \\
		\label{eq:loss}
	\end{split}
\end{equation}
\vspace{-0.7cm}
\begin{equation}
	\hat{x} = g_s\left(g_{eb}\left(g_a\left(x\right)\right)\right)
	\label{eq:functional_representation}
\end{equation}
where, $\mathcal{R}$ represents rate, $\lambda_1 \text{ and } \lambda_2$ are Lagrangian multipliers, and $\Psi$ indicates factorized prior. We compute the mean square error ($MSE$) and the multi-scale SSIM ($\mathit{MS\mbox{-}SSIM}$) between the original and reconstructed images as distortion measures. In our experiments, we empirically found $\lambda_1$ to be $\{0.01, 0.005, 0.0025, 0.00125\}$ and $\lambda_2$ to be $\{2.4, 1.2, 0.6, 0.3\}$ culminating in a bitrate between 0.1 and 0.4 bpp. For optimization we used Adam\cite{kingma2014adam} optimizer with an initial learning rate of $10^{-4}$.  Each model corresponding to a $\lambda$ value is trained for 450 epochs. 
\begin{figure}[t!]
	\centering
	\includegraphics[width=\linewidth]{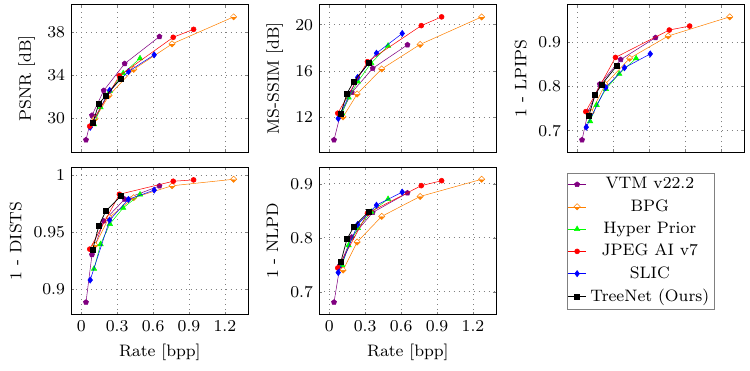}
	\vspace{-0.5cm}
	\caption{RD curves on CLIC Professional Valid dataset\cite{toderici2020workshop}.}
	\label{fig:rdplot_clic}
	\vspace{-0.3cm}
\end{figure}
\begin{figure}[t!]
	\centering
	\includegraphics[width=\linewidth]{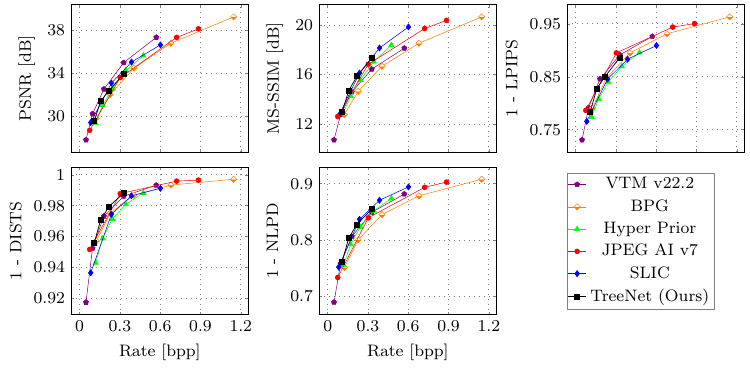}
	\vspace{-0.5cm}
	\caption{RD curves on Tecnick dataset\cite{asuni2014testimages}.}
	\label{fig:rdplot_technick}
	\vspace{-0.6cm}
\end{figure}
\begin{figure}[t!]
	\begin{center}
		\includegraphics[width=\linewidth]{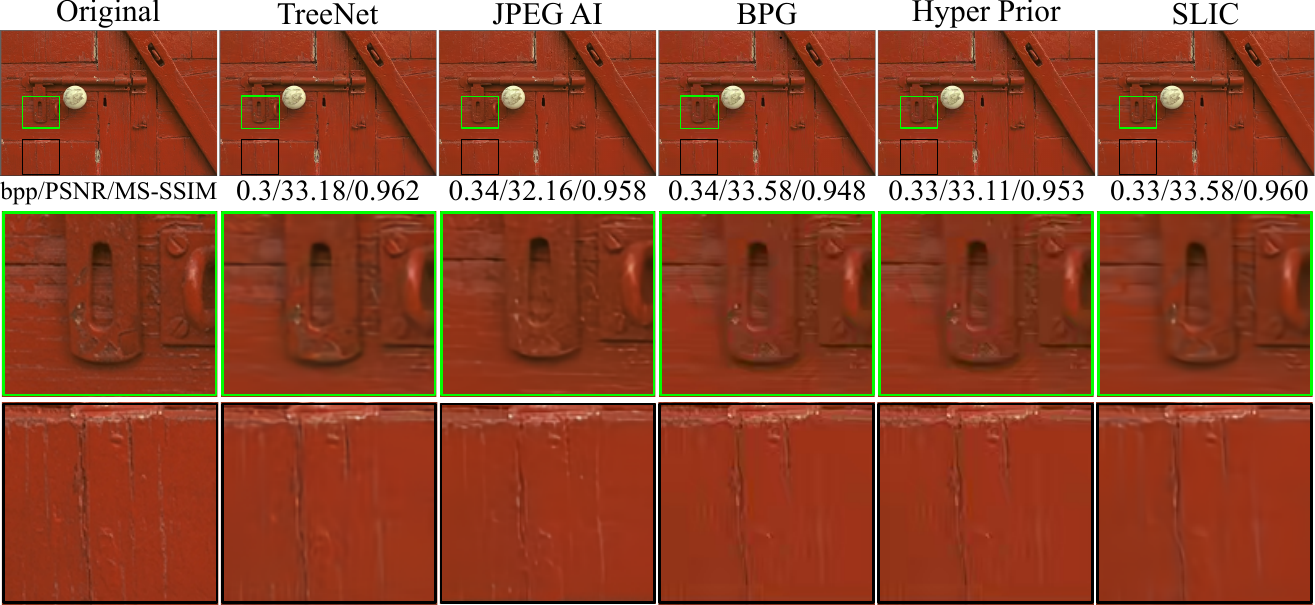}
	\end{center}
	\caption{Visual comparison of reconstructed images and enlarged patches on \emph{kodim02} image from Kodak dataset \cite{kodak1993}.}
	\label{fig:kodim_02}
	\vspace{-0.5cm}
\end{figure}
\subsection{Testing Setup}
\label{subsec:testing_setup}
For evaluation, we used Kodak\cite{kodak1993}, CLIC Professional Valid\cite{toderici2020workshop}, and Tecnick\cite{asuni2014testimages} datasets. 
We benchmarked TreeNet against BPG\cite{sullivan2012overview}, VTM \cite{bross2021overview}, JPEG AI\cite{jpegai}, Factorized Prior\cite{balle2017endtoend}, Hyper Prior\cite{ballé2018variational}, Color Learning\cite{color2023} and SLIC\cite{slic2024}. For quantitative comparison, we computed quality metrics such as PSNR, MS-SSIM\cite{wang2003multiscale}, LPIPS\cite{zhang2018unreasonable}, and DISTS\cite{ding2020image}  using PyTorch Image Quality package\cite{kastryulin2022piq}. Additionally, we compute NLPD\cite{laparra2016perceptual} using the implementation provided in \cite{ding2020optim}. We also calculate Bj{\o}ntegaard delta bitrate (BD-rate)\cite{bjontegaard2001calculation} for comparing different codecs.
\vspace{-0.2cm}
\section{Results}
\label{sec:result}
\begin{table*}[t]
	\centering
	\scriptsize
	\caption{BD-rate (\%) comparison. \textcolor{red}{\textbf{Red}} represents the best performance and \textcolor{blue}{\underline{\textbf{blue}}} indicates the second best performance.} 
	\begin{adjustbox}{width=\textwidth}
	\setlength{\tabcolsep}{1.5mm}{
		\begin{tabular}{@{}cccccccccccccc@{}}
			\toprule
			\multicolumn{1}{c|}{\multirow{2}{*}{\textbf{Methods}}}                            & \multicolumn{4}{c|}{\textbf{Kodak}\cite{kodak1993}}    & \multicolumn{4}{c|}{\textbf{CLIC Professional Valid}\cite{toderici2020workshop}}  & \multicolumn{4}{c}{\textbf{Tecnick}\cite{asuni2014testimages}}  \\
			\multicolumn{1}{c|}{}                                                     & \multicolumn{1}{c}{PSNR} & \multicolumn{1}{c}{MS-SSIM}& \multicolumn{1}{c}{1$-$NLPD} & \multicolumn{1}{c|}{1$-$LPIPS}& \multicolumn{1}{c}{PSNR} & \multicolumn{1}{c}{MS-SSIM}& \multicolumn{1}{c}{1$-$NLPD} & \multicolumn{1}{c|}{1$-$LPIPS}& \multicolumn{1}{c}{PSNR} & \multicolumn{1}{c}{MS-SSIM}& \multicolumn{1}{c}{1$-$NLPD} & 1$-$LPIPS \\ 		
		    \midrule
			\multicolumn{1}{c|}{BPG\cite{sullivan2012overview}}  & 0 & 0 & 0 & 	\multicolumn{1}{c|}{0} & 0  & 0    & 0   & \multicolumn{1}{c|}{0}   & 0   & 0   & 0   & 0     \\ 
			\multicolumn{1}{c|}{VTM\cite{bross2021overview}}  & \textcolor{red}{\textbf{-22.62}} & -15.01 & -23.52 & 	\multicolumn{1}{c|}{ \textcolor{red}{\textbf{-10.93}}} & \textcolor{red}{\textbf{-27.98}}  &  -20.20    & -27.46   & \multicolumn{1}{c|}{\textcolor{blue}{\underline{\textbf{-17.39}}}}  & \textcolor{red}{\textbf{-28.93}}   & -16.27   & -25.50   & \textcolor{red}{\textbf{-17.65}}   \\ 
			\multicolumn{1}{c|}{Color Learning\cite{color2023}}  & 49.03 & -7.02 & -6.59  & \multicolumn{1}{c|}{31.60} & 41.28  & -7.63 & -3.55  & \multicolumn{1}{c|}{22.10}  & 45.55 & 1.23 & 7.26 & 33.86  \\
			\multicolumn{1}{c|}{Factorized Prior\cite{balle2017endtoend}}  & 23.88 & -14.59 & -15.97  & \multicolumn{1}{c|}{30.22} & 24.01  & -16.77   & -11.04 & \multicolumn{1}{c|}{39.42} & 21.13 & -7.33 & -1.63 & 37.71 \\ 
			\multicolumn{1}{c|}{Hyper Prior\cite{ballé2018variational}}  & -2.39 & -18.80 & -23.18  & \multicolumn{1}{c|}{19.84} & -7.60  & -26.79   & -26.07  & \multicolumn{1}{c|}{15.27} & -5.88 & -19.16 & -18.51 &15.99\\ 		
			\multicolumn{1}{c|}{SLIC \cite{slic2024}}  &\textcolor{blue}{\underline{\textbf{-13.67}}} & -30.31 & \textcolor{red}{\textbf{-35.65}}  & \multicolumn{1}{c|}{9.23} & -8.65  & -34.05 & -31.80   & \multicolumn{1}{c|}{12.81} & \textcolor{blue}{\underline{\textbf{-18.37}}}& \textcolor{red}{\textbf{-33.07}} & \textcolor{red}{\textbf{-33.46}} & 7.77 \\ 				
			\multicolumn{1}{c|}{JPEG AI \cite{jpegai}} & 7.47 & \textcolor{red}{\textbf{-37.51}} & -14.36  & \multicolumn{1}{c|}{2.12} &\textcolor{blue}{\underline{\textbf{-15.06}}}  & \textcolor{red}{\textbf{-38.61}}  & \textcolor{blue}{\underline{\textbf{-33.11}}}  & \multicolumn{1}{c|}{\textcolor{red}{\textbf{-24.71}}} & -8.30 & -29.08 & -17.62 & \textcolor{blue}{\underline{\textbf{-17.09}}} \\
			\multicolumn{1}{c|}{TreeNet (Ours)}  & -2.95 & \textcolor{blue}{\underline{\textbf{-32.15}}} & \textcolor{blue}{ \underline{\textbf{-33.30}}}  & \multicolumn{1}{c|}{\textcolor{blue}{\underline{\textbf{-10.18}}}} & -13.34  & \textcolor{blue}{\underline{\textbf{-35.05}}} & \textcolor{red}{\textbf{-38.17}}  & \multicolumn{1}{c|}{-13.72} & -14.08 & \textcolor{blue}{\underline{\textbf{-30.73}}} & \textcolor{blue}{\underline{\textbf{-31.31}}} & -10.52 \\ 							
		\bottomrule			
	\end{tabular}}
	\label{tab:bd}
	\end{adjustbox}
	\vspace{-0.4cm}
\end{table*}
\subsection{Quantitative Comparison}
\label{subsec:quantitative_comparison}
For evaluating the performance of TreeNet quantitatively, we compute the rate-distortion (RD) performance on datasets as mentioned in Subsection~\ref{subsec:testing_setup}. The BD-rate computations are reported in Table~\ref{tab:bd} and the RD plots are shown in Fig.~\ref{fig:rdplot_kodak},~\ref{fig:rdplot_clic}, and~\ref{fig:rdplot_technick}.  Overall, TreeNet outperforms Factorized Prior\cite{balle2017endtoend}, Color Learning\cite{color2023}, Hyper Prior\cite{ballé2018variational}, and BPG\cite{sullivan2012overview}, while performing competitively compared to VTM\cite{bross2021overview}, JPEG AI\cite{jpegai}, and SLIC\cite{slic2024} across all metrics. TreeNet has an average BD-rate gain in PSNR of 4.83\% over JPEG AI. Notably, TreeNet has BD-rate savings of 12.50\%  in NLPD metric over JPEG AI\cite{jpegai} while being significantly less complex. 
\vspace{-0.2cm}
\subsection{Qualitative Comparison}
\vspace{-0.08cm}
\label{subsec:quanlitative_comparison}
To showcase the efficacy of TreeNet to produce high quality reconstructions, we visually compare the output of various methods to that of TreeNet as shown in Fig.~\ref{fig:kodim_02}. Along with the overall comparison, we focus on specific areas of the image to highlight the differences. Upon closer inspection, we observe TreeNet has higher reconstruction fidelity both in terms of structure and color compared to other codecs. Even though structural fidelity of JPEG AI\cite{jpegai} is better compared to our method, TreeNet outperforms it in terms of color fidelity. We observe that subtle changes in color are well preserved in TreeNet. 
\vspace{-0.15cm}
\subsection{Complexity Analysis}
\label{subsec:complexity_anaysis}
For quantifying complexity, we compute thousand multiply-accumulate operations per pixel (kMACs/pixel) using torchinfo\footnote{https://github.com/TylerYep/torchinfo} module. Table~\ref{tab:complexity_comparison} depicts a comparison of kMACs/pixel of various learning-based models. The overall complexity for TreeNet encompassing both encoder and decoder is 60.4 kMACs/pixel. Out of this, the decoder accounts for 51.08 kMACs/pixel, while the encoder contributes 9.32 kMACs/pixel. Notably, TreeNet has 87.82 \% less complexity compared to JPEG AI (495.99 kMACs/pixel). We further compute module-wise complexities along with the number of parameters  for TreeNet and report them in Table~\ref{tab:complexity}. 
\begin{table}
	\begin{center}
		\caption{Complexity comparison of various codecs.}
		\label{tab:complexity_comparison}
		\begin{tabular}{r|c|c}
			\toprule
			\textbf{Codec Names }& \begin{tabular}{@{}c@{}}\textbf{Encoder Complexity} \\ \textbf{{[}kMACs/pixel{]}}\end{tabular}& 
			\begin{tabular}{@{}c@{}}\textbf{Decoder Complexity} \\ \textbf{{[}kMACs/pixel{]}}\end{tabular}\\
			\midrule
			TreeNet (ours) & 9.32 & 51.08 \\
			Factorized Prior& 36.84 & 147.25 \\
			Hyper Prior & 40.80 & 149.89 \\
			JPEG AI & 277.16 & 218.83 \\
			Color Learning & 128.89 & 305.58 \\
			SLIC & 93.52 & 745.57\\
			\bottomrule
		\end{tabular}
	\end{center}
	\vspace{-0.3cm}
\end{table}
\begin{table}
	\begin{center}
		\caption{Complexity analysis of TreeNet.}
		\label{tab:complexity}
		\begin{tabular}{r|c|c}
			\toprule
			\textbf{Module Names} & \textbf{Complexity [kMACs/pixel]} & 
			\begin{tabular}{@{}c@{}}\textbf{No. Params. }\\ \textbf{{[}millions{]}}\end{tabular}\\
			\midrule
			$g_a$ & 6.86 & 0.32 \\
			$g_s$ & 49.89 & 0.86 \\
			$h_a$ $(\times 4)$ & 0.37 & 0.19 \\
			$h_s$ $(\times 4)$ & 0.85 & 0.68 \\
			$h_{ep}$ $(\times 4)$ & 0.8 & 0.03 \\
			Context Model $(\times 4)$ & 0.44 & 0.21\\
			\bottomrule
		\end{tabular}
	\end{center}
	\vspace{-0.5cm}
\end{table}
\vspace{-0.15cm}
\subsection{Ablation Study}
\subsubsection{Latent Interpretation}
For showcasing the impact of four input feature maps of $g_s$ on the reconstructed image, we conducted eight experiments belonging to two broad categories, namely, selective propagation and accumulative propagation. 

\begin{figure}[t!]
	\begin{center}
		\includegraphics[width=\linewidth]{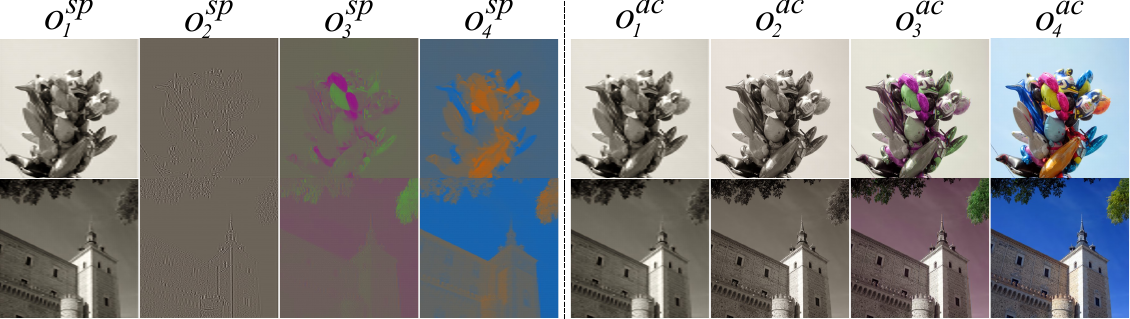}
	\end{center}
	\vspace{-0.1cm}
	\caption{Visualization of latent interpretation \emph{RGB\_OR\_1200$\times$1200\_014} and \emph{RGB\_OR\_1200$\times$1200\_005} images from Tecnick dataset \cite{asuni2014testimages}. }
	\label{fig:ablation_block_based}
	\vspace{-0.4cm}
\end{figure}

In selective propagation, we provide a single input feature map out of the four to $g_s$ at a time as shown in (\ref{eq:selective_propagation}) and inspect the output.  In doing so, we determine the influence of each feature map in the pixel space.
\begin{equation}
	\small
	o_i^{sp} = g_s\left(y_i\right);\  i \in \{1, 2, 3, 4\}
	\label{eq:selective_propagation}
\end{equation}
The first four columns in Fig.~\ref{fig:ablation_block_based}, depict the output $o_i^{sp}$ when individual input feature maps are provided to $g_s$. From these columns we can infer that $y_1$ and $y_2$ are responsible for reconstruction of low frequency and high frequency contents in the image, whereas $y_3$ and $y_4$ impart color to the reconstruction.
Secondly, in accumulative propagation, we gradually accumulate the input to $g_s$ one after the other, starting from the feature map $y_1$ as shown in (\ref{eq:ac}).
\begin{equation}
	\small
	\begin{split}
		o_i^{ac} &= g_s\left(y^{ac}_i\right); i \in \{1, 2, 3, 4\};\\
		y^{ac}_i &\in \{(y_1), (y_1, y_2), (y_1, y_2, y_3), (y_1, y_2, y_3, y_4)\}
	\end{split}
	 \label{eq:ac}
\end{equation}
We showcase the effect of such accumulative propagation in pixel space in the last four columns of Fig.~\ref{fig:ablation_block_based}. We observe that  providing $y_1$ as input to $g_s$ produces a low frequency output image. Providing $y_1$ and $y_2$ together results in the luma component of the image. As we accumulate $y_3$ and $y_4$ feature maps in the input, color components get added to the reconstructed image. Note that TreeNet decomposes the features into luma and chroma components without being directly supervised to do so during training.
\vspace{0.1cm}
\subsubsection{Spatial Rate Distribution}
We visualize the average number of bits required for encoding the latent feature maps. The bitmap is computed by averaging likelihoods for latent pixels across channels.  Formally, the bitmap generation process can be stated as
\begin{equation}
	\small
	{M}_{y_i} = \Omega\left(-\frac{1}{C}\sum_{j = 0}^{C-1}\text{log}_\text{2}\left(\mathcal{E}_{\hat{y}_{i_{j}}|\hat{z}_{i_{j}}}\left(\hat{y}_{i_{j}}|\hat{z}_{i_{j}}\right)\right)\right)
	\label{eq:bitmap}
\end{equation}
where ${M}_{y_i}$ represents the upscaled bitmap used for visualization.  $\Omega(\cdot)$ is the nearest neighbour operation used for scaling the bitmap to image dimension, $C$ is the number of channels in a latent feature map, $\hat{y}_{i_{j}}$ indicates the $j^{th}$ channel of $i^{th}$ quantized latent $\hat{y}$, and $\hat{z}_{i_{j}}$ represents the $j^{th}$ channel of $i^{th}$ quantized hyper-latent $\hat{z}$.

\begin{figure}
	\begin{center}
		  \includegraphics[width=\linewidth]{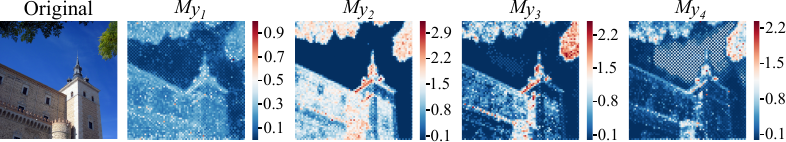}
	\end{center}
	\vspace{-0.3cm}
	\caption{Bitmaps for 
	\emph{RGB\_OR\_1200$\times$1200\_005} image from Tecnick dataset \cite{asuni2014testimages}. }
	\label{fig:ablation_bits_vis}
	\vspace{-0.65cm}
\end{figure}

We visualize the bitmaps alongside the original image in Fig.~\ref{fig:ablation_bits_vis}. 
For latent $y_1$, the bitmap is more spread out compared to that for latent $y_2$ for which the bitmap is concentrated around the high frequency parts of the image. 
The bitmaps for latents $y_3$ and $y_4$ present the focus on color gradient. The checkerboard patterns that are visible in bitmaps are due to the checkerboard context model.
\vspace{-0.1cm}

\section{Conclusion}
\vspace{-0.1cm}
In this paper, we propose a novel learning-based image compression method called TreeNet that leverages a binary tree-structured architecture for complexity reduction. We present a detailed quantitative and qualitative evaluation of our method and compare it with various state-of-the-art methods including JPEG AI. The experiments showcase competitiveness of TreeNet across three test sets in the five evaluation metrics while being significantly less complex. Finally, we elucidate the contribution of latent blocks on reconstruction providing interpretability to our model. In future work, we aim to further reduce the decoder complexity and improve the rate-distortion performance through better context modeling.
\vspace{-0.2cm}
\bibliographystyle{IEEEtran}
{ \bibliography{refs}}

\begin{thebibliography}{10}
\providecommand{\url}[1]{#1}
\csname url@samestyle\endcsname
\providecommand{\newblock}{\relax}
\providecommand{\bibinfo}[2]{#2}
\providecommand{\BIBentrySTDinterwordspacing}{\spaceskip=0pt\relax}
\providecommand{\BIBentryALTinterwordstretchfactor}{4}
\providecommand{\BIBentryALTinterwordspacing}{\spaceskip=\fontdimen2\font plus
\BIBentryALTinterwordstretchfactor\fontdimen3\font minus
  \fontdimen4\font\relax}
\providecommand{\BIBforeignlanguage}[2]{{%
\expandafter\ifx\csname l@#1\endcsname\relax
\typeout{** WARNING: IEEEtran.bst: No hyphenation pattern has been}%
\typeout{** loaded for the language `#1'. Using the pattern for}%
\typeout{** the default language instead.}%
\else
\language=\csname l@#1\endcsname
\fi
#2}}
\providecommand{\BIBdecl}{\relax}
\BIBdecl

\bibitem{skodras2001jpeg}
A.~Skodras, C.~Christopoulos, and T.~Ebrahimi, ``The jpeg 2000 still image
  compression standard,'' \emph{IEEE Signal processing magazine}, vol.~18,
  no.~5, pp. 36--58, 2001.

\bibitem{sullivan2012overview}
G.~J. Sullivan, J.-R. Ohm, W.-J. Han, and T.~Wiegand, ``Overview of the high
  efficiency video coding (hevc) standard,'' \emph{IEEE Transactions on
  circuits and systems for video technology}, vol.~22, no.~12, pp. 1649--1668,
  2012.

\bibitem{bross2021overview}
B.~Bross, Y.-K. Wang, Y.~Ye, S.~Liu, J.~Chen, G.~J. Sullivan, and J.-R. Ohm,
  ``Overview of the versatile video coding (vvc) standard and its
  applications,'' \emph{IEEE Transactions on Circuits and Systems for Video
  Technology}, vol.~31, no.~10, pp. 3736--3764, 2021.

\bibitem{balle2017endtoend}
J.~Ball{\'e}, V.~Laparra, and E.~P. Simoncelli, ``End-to-end optimized image
  compression,'' in \emph{International Conference on Learning
  Representations}, 2017.

\bibitem{theis2017lossy}
L.~Theis, W.~Shi, A.~Cunningham, and F.~Husz{\'a}r, ``Lossy image compression
  with compressive autoencoders,'' in \emph{International Conference on
  Learning Representations}, 2017.

\bibitem{ballé2018variational}
J.~Ballé, D.~Minnen, S.~Singh, S.~J. Hwang, and N.~Johnston, ``Variational
  image compression with a scale hyperprior,'' in \emph{International
  Conference on Learning Representations}, 2018.

\bibitem{minen2018}
D.~Minnen, J.~Ball\'{e}, and G.~D. Toderici, ``Joint autoregressive and
  hierarchical priors for learned image compression,'' in \emph{Advances in
  Neural Information Processing Systems}, S.~Bengio, H.~Wallach, H.~Larochelle,
  K.~Grauman, N.~Cesa-Bianchi, and R.~Garnett, Eds., vol.~31.\hskip 1em plus
  0.5em minus 0.4em\relax Curran Associates, Inc., 2018.

\bibitem{color2023}
S.~Prativadibhayankaram, T.~Richter, H.~Sparenberg, and S.~Foessel, ``Color
  learning for image compression,'' in \emph{2023 IEEE International Conference
  on Image Processing (ICIP)}.\hskip 1em plus 0.5em minus 0.4em\relax IEEE,
  2023, pp. 2330--2334.

\bibitem{slic2024}
S.~Prativadibhayankaram, M.~P. Panda, T.~Richter, H.~Sparenberg,
  S.~F{\"o}{\ss}el, and A.~Kaup, ``Slic: a learned image codec using structure
  and color,'' in \emph{2024 Data Compression Conference (DCC)}.\hskip 1em plus
  0.5em minus 0.4em\relax IEEE, 2024, pp. 3--12.

\bibitem{cheng2020learned}
Z.~Cheng, H.~Sun, M.~Takeuchi, and J.~Katto, ``Learned image compression with
  discretized gaussian mixture likelihoods and attention modules,'' in
  \emph{Proceedings of the IEEE/CVF conference on computer vision and pattern
  recognition}, 2020, pp. 7939--7948.

\bibitem{he2022elic}
D.~He, Z.~Yang, W.~Peng, R.~Ma, H.~Qin, and Y.~Wang, ``Elic: Efficient learned
  image compression with unevenly grouped space-channel contextual adaptive
  coding,'' in \emph{Proceedings of the IEEE/CVF Conference on Computer Vision
  and Pattern Recognition}, 2022, pp. 5718--5727.

\bibitem{he2021checkerboard}
D.~He, Y.~Zheng, B.~Sun, Y.~Wang, and H.~Qin, ``Checkerboard context model for
  efficient learned image compression,'' in \emph{Proceedings of the IEEE/CVF
  Conference on Computer Vision and Pattern Recognition}, 2021, pp.
  14\,771--14\,780.

\bibitem{jiang2023mlic}
W.~Jiang, J.~Yang, Y.~Zhai, P.~Ning, F.~Gao, and R.~Wang, ``Mlic:
  Multi-reference entropy model for learned image compression,'' in
  \emph{Proceedings of the 31st ACM International Conference on Multimedia},
  2023, pp. 7618--7627.

\bibitem{jiang2023mlicpp}
W.~Jiang, J.~Yang, Y.~Zhai, F.~Gao, and R.~Wang, ``Mlic++: Linear complexity
  multi-reference entropy modeling for learned image compression,'' \emph{arXiv
  preprint arXiv:2307.15421}, 2023.

\bibitem{kingma2013auto}
D.~P. Kingma, M.~Welling \emph{et~al.}, ``Auto-encoding variational bayes,''
  2013.

\bibitem{jpegai}
E.~Alshina, J.~Ascenso, and T.~Ebrahimi, ``Jpeg ai: The first international
  standard for image coding based on an end-to-end learning-based approach,''
  \emph{IEEE MultiMedia}, vol.~31, no.~4, pp. 60--69, 2024.

\bibitem{asuni2014testimages}
N.~Asuni, A.~Giachetti \emph{et~al.}, ``Testimages: a large-scale archive for
  testing visual devices and basic image processing algorithms.'' in
  \emph{STAG}, 2014, pp. 63--70.

\bibitem{kodak1993}
E.~Kodak, ``Kodak lossless true color image suite,'' \emph{Tech. Rep}, 1993.

\bibitem{minnen2020channel}
D.~Minnen and S.~Singh, ``Channel-wise autoregressive entropy models for
  learned image compression,'' in \emph{2020 IEEE International Conference on
  Image Processing (ICIP)}.\hskip 1em plus 0.5em minus 0.4em\relax IEEE, 2020,
  pp. 3339--3343.

\bibitem{mentzer2023m2t}
F.~Mentzer, E.~Agustson, and M.~Tschannen, ``M2t: Masking transformers twice
  for faster decoding,'' in \emph{Proceedings of the IEEE/CVF International
  Conference on Computer Vision}, 2023, pp. 5340--5349.

\bibitem{zhu2022transformerbased}
Y.~Zhu, Y.~Yang, and T.~Cohen, ``Transformer-based transform coding,'' in
  \emph{International Conference on Learning Representations}, 2022.

\bibitem{yang2023computationally}
Y.~Yang and S.~Mandt, ``Computationally-efficient neural image compression with
  shallow decoders,'' in \emph{Proceedings of the IEEE/CVF International
  Conference on Computer Vision}, 2023, pp. 530--540.

\bibitem{galpinAssym}
F.~Galpin, M.~Balcilar, F.~Lefebvre, F.~Racapé, and P.~Hellier, ``Entropy
  coding improvement for low-complexity compressive auto-encoders,'' in
  \emph{2023 Data Compression Conference (DCC)}, 2023, pp. 338--338.

\bibitem{leguay2023low}
T.~Leguay, T.~Ladune, P.~Philippe, G.~Clare, F.~Henry, and O.~D{\'e}forges,
  ``Low-complexity overfitted neural image codec,'' in \emph{2023 IEEE 25th
  International Workshop on Multimedia Signal Processing (MMSP)}.\hskip 1em
  plus 0.5em minus 0.4em\relax IEEE, 2023, pp. 1--6.

\bibitem{kim2024c3}
H.~Kim, M.~Bauer, L.~Theis, J.~R. Schwarz, and E.~Dupont, ``C3:
  High-performance and low-complexity neural compression from a single image or
  video,'' in \emph{Proceedings of the IEEE/CVF Conference on Computer Vision
  and Pattern Recognition}, 2024, pp. 9347--9358.

\bibitem{blard2024overfitted}
T.~Blard, T.~Ladune, P.~Philippe, G.~Clare, X.~Jiang, and O.~D{\'e}forges,
  ``Overfitted image coding at reduced complexity,'' in \emph{2024 32nd
  European Signal Processing Conference (EUSIPCO)}.\hskip 1em plus 0.5em minus
  0.4em\relax IEEE, 2024, pp. 927--931.

\bibitem{flepp2024real}
R.~Flepp, A.~Ignatov, R.~Timofte, and L.~Van~Gool, ``Real-world mobile image
  denoising dataset with efficient baselines,'' in \emph{Proceedings of the
  IEEE/CVF Conference on Computer Vision and Pattern Recognition}, 2024, pp.
  22\,368--22\,377.

\bibitem{dai21aff}
Y.~Dai, F.~Gieseke, S.~Oehmcke, Y.~Wu, and K.~Barnard, ``Attentional feature
  fusion,'' in \emph{{IEEE} Winter Conference on Applications of Computer
  Vision, {WACV} 2021}, 2021.

\bibitem{lin2014microsoft}
T.-Y. Lin, M.~Maire, S.~Belongie, J.~Hays, P.~Perona, D.~Ramanan,
  P.~Doll{\'a}r, and C.~L. Zitnick, ``Microsoft coco: Common objects in
  context,'' in \emph{Computer vision--ECCV 2014: 13th European conference,
  zurich, Switzerland, September 6-12, 2014, proceedings, part v 13}.\hskip 1em
  plus 0.5em minus 0.4em\relax Springer, 2014, pp. 740--755.

\bibitem{kingma2014adam}
D.~P. Kingma and J.~Ba, ``Adam: A method for stochastic optimization,''
  \emph{arXiv preprint arXiv:1412.6980}, 2014.

\bibitem{toderici2020workshop}
G.~Toderici, W.~Shi, R.~Timofte, L.~Theis, J.~Balle, E.~Agustsson, N.~Johnston,
  and F.~Mentzer, ``Workshop and challenge on learned image compression
  (clic2020),'' in \emph{CVPR}, 2020.

\bibitem{wang2003multiscale}
Z.~Wang, E.~P. Simoncelli, and A.~C. Bovik, ``Multiscale structural similarity
  for image quality assessment,'' in \emph{The Thrity-Seventh Asilomar
  Conference on Signals, Systems \& Computers, 2003}, vol.~2.\hskip 1em plus
  0.5em minus 0.4em\relax Ieee, 2003, pp. 1398--1402.

\bibitem{zhang2018unreasonable}
R.~Zhang, P.~Isola, A.~A. Efros, E.~Shechtman, and O.~Wang, ``The unreasonable
  effectiveness of deep features as a perceptual metric,'' in \emph{Proceedings
  of the IEEE conference on computer vision and pattern recognition}, 2018, pp.
  586--595.

\bibitem{ding2020image}
K.~Ding, K.~Ma, S.~Wang, and E.~P. Simoncelli, ``Image quality assessment:
  Unifying structure and texture similarity,'' \emph{IEEE transactions on
  pattern analysis and machine intelligence}, vol.~44, no.~5, pp. 2567--2581,
  2020.

\bibitem{kastryulin2022piq}
S.~Kastryulin, J.~Zakirov, D.~Prokopenko, and D.~V. Dylov, ``Pytorch image
  quality: Metrics for image quality assessment,'' 2022.

\bibitem{laparra2016perceptual}
V.~Laparra, J.~Ball{\'e}, A.~Berardino, and E.~P. Simoncelli, ``Perceptual
  image quality assessment using a normalized laplacian pyramid,''
  \emph{Electronic Imaging}, vol.~28, pp. 1--6, 2016.

\bibitem{ding2020optim}
K.~Ding, K.~Ma, S.~Wang, and E.~P. Simoncelli, ``Comparison of image quality
  models for optimization of image processing systems,'' \emph{CoRR}, vol.
  abs/2005.01338, 2020.

\bibitem{bjontegaard2001calculation}
G.~Bjontegaard, ``Calculation of average psnr differences between rd-curves,''
  \emph{ITU SG16 Doc. VCEG-M33}, 2001.

\end{thebibliography}
\end{document}